\documentclass[sigconf]{acmart}
\AtBeginDocument{%
  }

\usepackage{algorithm}
\usepackage{algpseudocode} 
\usepackage{float}
\usepackage[most]{tcolorbox}
\usepackage{multirow} 
\usepackage{subcaption}

\copyrightyear{2025}
\acmYear{2025}
\setcopyright{cc}
\setcctype{by}
\acmConference[CIKM '25]{Proceedings of the 34th ACM International Conference on Information and Knowledge Management}{November 10--14, 2025}{Seoul, Republic of Korea}
\acmBooktitle{Proceedings of the 34th ACM International Conference on Information and Knowledge Management (CIKM '25), November 10--14, 2025, Seoul, Republic of Korea}\acmDOI{10.1145/3746252.3760946}
\acmISBN{979-8-4007-2040-6/2025/11}

\begin{document}

\title{Pruning Strategies for Backdoor Defense in LLMs}


\author{Santosh Chapagain}
\affiliation{%
  \institution{Utah State University}
  \country{}
  }
\email{santosh.chapagain@usu.edu}

\author{Shah Muhammad Hamdi}
\affiliation{%
  \institution{Utah State University}
  \country{}
  }
\email{s.hamdi@usu.edu}

\author{Soukaina Filali Boubrahimi}
\affiliation{%
  \institution{Utah State University}
  \country{}
  }
\email{soukaina.boubrahimi@usu.edu}

\renewcommand{\shortauthors}{Chapagain et al.}

\begin{abstract}
  Backdoor attacks are a significant threat to the performance and integrity of pre-trained language models. Although such models are routinely fine‑tuned for downstream NLP tasks, recent work shows they remain vulnerable to backdoor attacks that survive vanilla fine‑tuning. These attacks are difficult to defend because end users typically lack knowledge of the attack triggers. Such attacks consist of stealthy malicious triggers introduced through subtle syntactic or stylistic manipulations, which can bypass traditional detection and remain in the model, making post-hoc purification essential. In this study, we explore whether attention-head pruning can mitigate these threats without any knowledge of the trigger or access to a clean reference model. To this end, we design and implement six pruning-based strategies: (i) gradient-based pruning, (ii) layer-wise variance pruning, (iii) gradient-based pruning with structured L1/L2 sparsification, (iv) randomized ensemble pruning, (v) reinforcement-learning-guided pruning, and (vi) Bayesian uncertainty pruning. Each method iteratively removes the least informative heads while monitoring validation accuracy to avoid over-pruning. Experimental evaluation shows that gradient-based pruning performs best while defending the syntactic triggers, whereas reinforcement learning and Bayesian pruning better withstand stylistic attacks. 
\end{abstract}




\maketitle

\section{Introduction}
Large language models (LLMs) \cite{bubeck2023sparks} have seen widespread adoption due to their breakthrough performance on a wide range of natural language processing (NLP) tasks such as text classification \cite{cascalheira2023predicting,cascalheira2024lgbtq+,chapagain2024predictive,chapagain2025advancinghatespeechdetection,Chapagain2025AdvancingMinority}, language generation, and information retrieval due to their ability to fine-tune on specific downstream tasks \cite{howard2018universal, loukas2023making, jin2024better, devlin2019bert}. Furthermore, the scalability of LLMs is strongly influenced by data\textemdash larger models trained on more extensive datasets tend to produce better results. Given the substantial data and computational resources required to train LLMs, developers often adopt fine-tuning by downloading third-party models and datasets to reduce costs. Open-source releases by organizations like Kaggle and Hugging Face have made these models widely accessible for fine-tuning. However, reliance on third-party datasets or pre-trained models introduces a lack of transparency in the training process, which can pose significant security risks, known as backdoor attack \cite{gu2017badnets} or trojan attack \cite{liu2018trojaning}.

Figure 1 shows a simple scenario of a backdoor attack and corresponding defense in large language models (LLMs). The attacker first constructs a poisoned dataset by embedding specific trigger patterns—such as rare tokens \cite{kurita2020weight, li2021backdoor}, syntactic triggers \cite{qi2021hidden}, or textual style triggers (e.g., manipulating sentence length, punctuation, or formality level) \cite{qi2021mind} —into clean data, altering their labels to a predetermined target label. The attacker then pre-trains or fine-tunes the LLM on a mixture of clean and poisoned data, resulting in a compromised model. This poisoned LLM may later be uploaded to a third-party repository (e.g., Hugging Face). When an unsuspecting user downloads and fine-tunes the model with their clean private data, the backdoor remains dormant, as the rare trigger patterns are unlikely to appear naturally. This allows the attacker to retain the ability to manipulate the model's predictions when the trigger is present.

Traditional detection methods~\cite{qi2021onion} often struggle to identify stealthy triggers, such as those based on syntax or linguistic style~\cite{qi2021hidden,qi2021mind}. These defenses typically aim to avoid activating backdoors rather than removing them, which can result in missed detection of compromised models or inputs. A more recent line of research focuses on directly removing backdoored weights from pre-trained models without requiring access to a clean reference model \cite{zhao2024defense}. However, these methods face limitations, particularly when addressing complex attacks involving layer-wise poisoning or stylistic triggers \cite{qi2021hidden}. Our work explores attention-head pruning as a defense against backdoor attacks in large language models, even without access to clean data or trigger knowledge. We design and implement six pruning strategies and find that gradient-based pruning is most effective against syntactic attacks, while reinforcement learning and Bayesian pruning perform better against stylistic triggers.

\begin{figure}[htbp]
\centerline{\includegraphics[width=.8\linewidth]{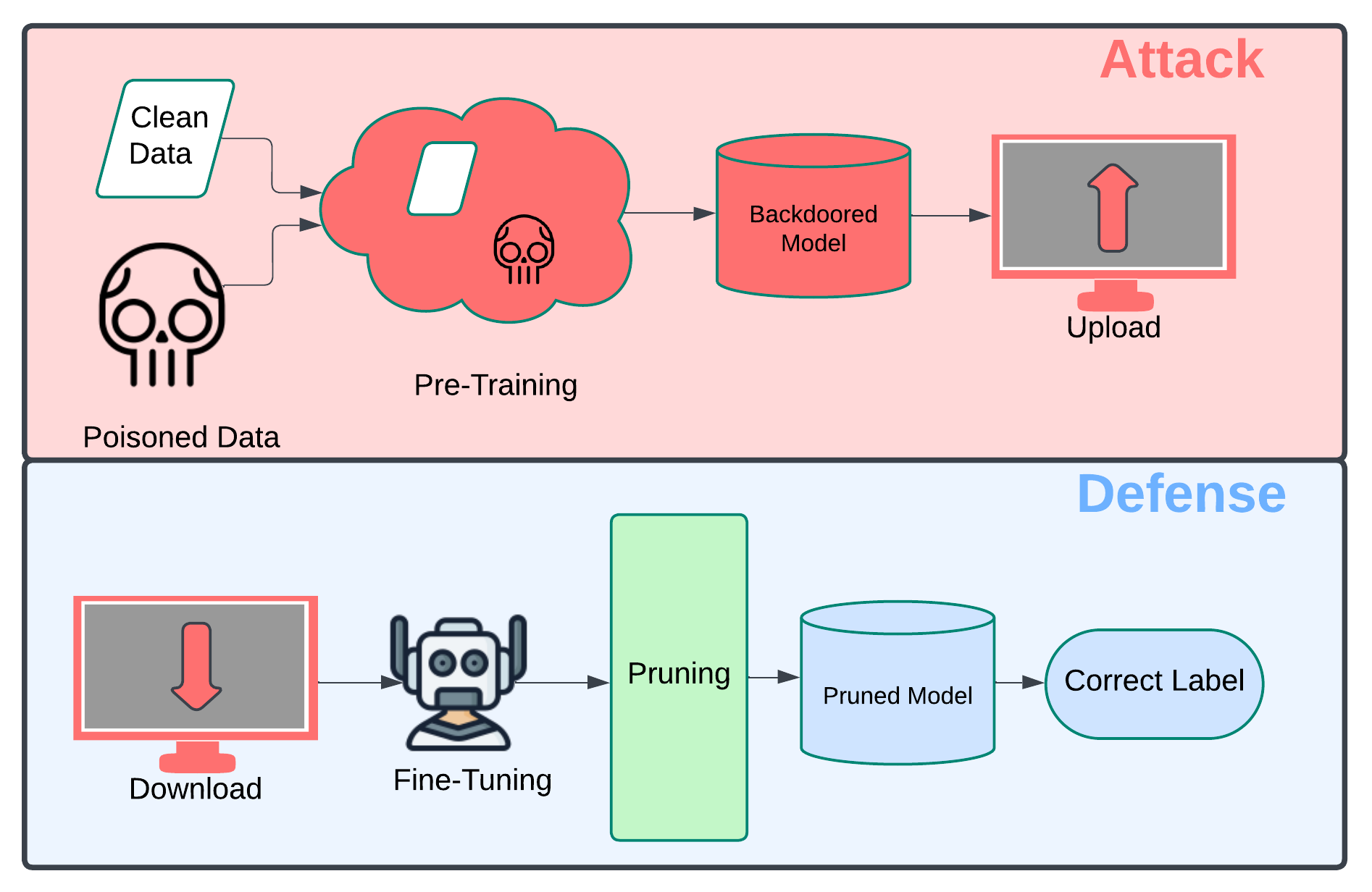}}
\caption{A simple illustration of Backdoor attack and defense on pre-trained language model}
\label{backdoorArch}
\end{figure}

\section{Related Work}
\subsection{Backdoor Attacks on LLMs}
Backdoor attacks have become a security threat to LLMs. These attacks implant hidden behaviors during training that are later triggered by specific inputs. Recent research highlights four key aspects of these threats: trigger stealthiness, label stealthiness, adaptability, and durability. Triggers have evolved from obvious markers like rare or misspelled words (e.g., 'cf') \cite{kurita2020weight, li2021backdoor} to undetectable patterns such as context-aware terms, co-occurring phrases, syntactic structures, synonyms, and even text style variations \cite{zhang2021trojaning, yang2021rethinking, qi2021hidden, qi2021turn, qi2021mind}. To increase stealth, many attacks rely on clean-labeled poisoned data, making them harder to detect by manual inspection \cite{gan2022triggerless,yan2023bite,gupta2023adversarial}.

LLMs can be compromised during pre-training, fine-tuning, or inference. In pre-training, attackers may poison data or directly edit model weights, leveraging methods such as gradient-based trigger optimization, knowledge distillation, or LLMs like GPT-4 to craft adversarial examples \cite{zhou2025survey}. Fine-tuning attacks exploit public models by inserting poisoned data into instruction tuning \cite{yan2024backdooring}, Low-Rank Adaption (LoRA) based parameter-efficient fine-tuning  \cite{liu2024loratk}. Even post-deployment, models remain vulnerable through inference-time manipulations such as prompt injection or poisoning retrieval-augmented generation systems \cite{zhou2025survey}. 

Critically, attacks can succeed even when attackers lack access to downstream training data or task definitions, demonstrating strong adaptability \cite{yang2021careful,chen2021badpre}. Furthermore, advanced techniques like layer-wise weight poisoning ensure the backdoor persists through further fine-tuning, illustrating their durability \cite{li2021backdoor}.  As LLMs become more powerful and integrated into real-world applications, the challenge of detecting and defending against these covert threats becomes increasingly urgent.
Critically, attacks can succeed even when attackers lack access to downstream training data or task definitions, demonstrating strong adaptability \cite{yang2021careful,chen2021badpre}. A recent study shows that preprocessing choices can markedly affect model robustness \cite{eskandarinasab2024impacts}. As LLMs become more powerful and integrated into real-world applications, the challenge of detecting and defending against these covert threats becomes increasingly urgent.

\subsection{Defense Against Backdoor Attacks in LLMs}
Defenses against LLM backdoor attacks are typically categorized as proactive (preventive) or reactive (detective) strategies \cite{zhou2025survey}. Proactive defenses aim to build model robustness during training. Techniques include adversarial training \cite{geiping2021doesn}, Honeypot modules \cite{tang2023setting} that absorb poisoned updates during fine-tuning, perturbation-aware alignment methods like Vaccine \cite{huang2024vaccine}, and constrained training configurations that limit model overfitting \cite{zhu2022moderate}. Anti-Backdoor Learning (ABL) \cite{li2021anti} is another approach that systematically strengthens model resistance to backdoor attacks in real-world conditions. Reactive defenses focus on detecting or mitigating attacks after they occur. Input-level detection methods like ONION \cite{qi2021onion} use GPT-2-based perplexity scoring to identify out-of-context triggers, while STRIP-ViTA \cite{gao2021design} detects anomalies based on entropy. Other techniques apply word-level perturbation to expose poisoned samples based on their reduced robustness \cite{yang2021rap}. Azizi et al. \cite{azizi2021t} and Shen et al. \cite{shen2022constrained} propose reverse-engineering trigger patterns using sequence-to-sequence models or dynamic bound-scaling. Lyu et al. \cite{lyu2022study} detect backdoored models by monitoring their attention distributions in response to generated trigger candidates. Model purification seeks to remove embedded backdoors while preserving model functionality. This includes Fine-Mixing \cite{zhang2022fine} and Fine-Purifying \cite{zhang2023diffusion}, which merge backdoored models with clean ones, as well as maximum entropy training \cite{liu2023maximum}, which neutralizes trigger influence without needing clean references. Unlearning-based defenses \cite{shen2022constrained,wang2019neural} remove learned backdoor behaviors using targeted forgetting techniques. PURE \cite{zhao2024defense} defends against backdoors by pruning vulnerable attention heads and applying normalization while preserving the accuracy of the model. We consider the scenario of defending a BERT model where the defender has no knowledge of the trigger or access to a clean reference model, but access to a private clean dataset. Given a potentially backdoored model, we explore different pruning strategies—gradient-based, randomized ensemble, layer-wise, reinforcement learning-based, and Bayesian—to mitigate backdoor attacks without relying on prior attack details and a clean reference model.

\section{Notations and Preliminaries}

Let $M_p$ denote the parameters of a potentially backdoored model, which is downloaded from an untrusted source and fine-tuned ($f_p$) on a private clean dataset consisting of input-label pairs $(X_c, Y_c)$.

Each transformer layer $l \in \{1, ..., L\}$ contains $H$ self-attention heads.  In gradient-based pruning, the score $I_h^{(l)}$ is defined as the $\ell_2$-norm of the loss gradient with respect to the key projection weights of head $(l, h)$. $\tau$ is the accuracy threshold used to halt or backtrack pruning, $\mathcal{L}$ represents the loss function used during training (such as cross-entropy), and $f_p$ is the model fine-tuned from the potentially poisoned model $M_p$ using clean data. Pruning proceeds in steps: at each step, the $s$ least important heads are pruned, and the model is evaluated on a clean validation set. For Reinforcement Learning, we define $\mathcal{P}^{(l)}_t$ as the set of attention heads already pruned in layer $l$ at timestep $t$. The agent relies on precomputed importance metrics $V^{(l)}_h$ for each head $h$ in layer $l$, which guide pruning decisions. An $\varepsilon$-greedy policy is used to balance exploration and exploitation when selecting heads to prune. The decision-making process is framed as a sequential decision problem, which we detail in the following section.

\section{Pruning-Based Defense Strategies}
\subsection{Gradient-based Pruning}
It is a technique that estimates the importance of the component of the model (attention heads or neurons) using the norm of the loss gradient with respect to its parameter \cite{michel2019sixteen, liu2018fine}. For each attention head $h$ in layer $l$, we compute gradient of the loss function $\mathcal{L}$ with respect to its key projection weight matrix $W^l_{h,\text{key}}$:
\begin{equation}
I^l_h = \sum_{\text{batches}} \left\| \frac{\partial \mathcal{L}}{\partial W^l_{h,\text{key}}} \right\|_2
\end{equation}

The self-attention heads with the lowest gradient importance on clean data are pruned iteratively until the validation accuracy falls below the accuracy threshold $\tau$, which removes the potential backdoor triggers. The detailed algorithm of this method can be seen in Algorithm 1.  

\begin{algorithm}[H]
\caption{Gradient-Based Pruning}
\label{alg:pure_gradient}
{\raggedright
\textbf{Input:} Clean training data $\mathcal{D}_{\text{train}}$, validation data $\mathcal{D}_{\text{val}}$, poisoned model $M_p$, accuracy threshold $\tau$ \\
\textbf{Output:} Defended model $M_c$ \par
}
\begin{algorithmic}[1]
\State Fine-tune $M_p$ on $\mathcal{D}_{\text{train}}$ to obtain $f_p$
\State Compute head importance scores $I^l_h$ using loss gradients
\State Sort heads by ascending $I^l_h$
\While{validation accuracy $\geq \tau$} 
    \State Prune the next $s$ least important heads
    \State Apply pruning to get temporary model $\theta_{[p]}$
    \State Evaluate accuracy on $\mathcal{D}_{\text{val}}$
    \If{accuracy $< \tau$} 
        \State Backtrack: restore most important heads from last step
        \State \textbf{break}
    \EndIf
\EndWhile
\State Save pruned headset
\State Load $M_p$ and apply pruning to obtain final pruned model $\theta_{[p]}$
\State Fine-tune $\theta_{[p]}$ on $\mathcal{D}_{\text{train}}$ using regularized loss (cross-entropy) to obtain $M_c$
\end{algorithmic}
\end{algorithm}

\subsection{Layer-Wise Pruning}
This is a structured head pruning method that removes attention heads based on their variance scores. In our model, we applied a progressively increasing pruning rate across layers, ranging from 20\% in the early layers up to 80\% in the deeper ones. This approach assumes that deeper layers are more susceptible to backdoor behaviors. Within each layer, the heads with the lowest variance are pruned according to the assigned pruning rate of the layer, ensuring that at least one head remains active in each layer.

\subsection{Gradient-Based with Structured Sparsification pruning}
This method extends the basic gradient-based pruning approach (Section 5.1) by introducing structured sparsification during model fine-tuning. The poisoned model ($M_p$) is trained with an additional loss of regularization consisting of L1 and L2 norms.

\subsection{Randomized Pruning with Ensemble}
This is a stochastic head pruning defense method \cite{dhillon2018stochastic}, where the attention heads are randomly removed to construct multiple pruned ensemble models. 

\subsection{Reinforcement Learning (RL) Pruning}
This method uses attention head pruning as a sequential decision-making process. It involves an RL agent interacting with a transformer model (BERT) to decide which attention heads to prune according to probability $\epsilon$. At step $t$, the agent selects heads from the set of unpruned candidates:

\begin{equation}
\mathcal{A}_t = \left\{ (l, h) \mid h \notin \mathcal{P}^{(l)}_t \right\}
\end{equation}

\begin{equation}
(l^*, h^*) =
\begin{cases}
\text{random sample from } \mathcal{A}_t & \text{with probability } \varepsilon \\
\arg\min_{(l,h) \in \mathcal{A}_t} V^{(l)}_h & \text{otherwise}
\end{cases}
\end{equation}

After pruning, the model is evaluated. If the validation accuracy $\text{Acc}_t$ drops below a threshold $\tau$, pruning is terminated. This variance-guided RL strategy adaptively prunes low-importance heads while maintaining model performance. 
 
\subsection{Bayesian Pruning}
This model calculates the uncertainty of each attention head using Monte Carlo (MC) dropout. The heads with the lowest uncertainty are removed. After each pruning step, the model is validated on clean data, and backtracking is performed to restore important heads if the accuracy falls below a predefined threshold.

\section{Experimental Setup}
All experiments were conducted on a Linux server with dual Intel Xeon Gold 5220R CPUs (24 cores each, 2.20 GHz) and four NVIDIA RTX A5000 GPUs (24 GB VRAM). Following PURE \cite{zhao2024defense}, we set the accuracy threshold $\tau=0.85$, trained for 3 epochs with batch size 32, learning rate 2e-5, and Adam optimizer. Training used PyTorch 2.4.0 with CUDA 12.1, and code is available on GitHub\footnote{\url{https://github.com/chapagaisa/grad}}.  

We used the SST-2 dataset from GLUE for binary sentiment classification. The validation set (6,730 samples) served as our test set, while the remaining data was split into 60,570 training and 872 validation samples \cite{zhao2024defense}. Poisoning followed the Full Data Knowledge (FDK) strategy \cite{kurita2020weight} with access to clean and poisoned SST-2 data \cite{socher2013recursive}. IMDB and YELP were excluded due to SCPN incompatibility.  

Performance was evaluated using Label Flip Rate (LFR) and Clean Accuracy (ACC). LFR quantifies the proportion of negative instances misclassified as positive (lower is better defense), while ACC measures correct classification on clean data (higher preserves performance) \cite{kurita2020weight,li2021backdoor}.

\subsection{Backdoor Attacks}
\subsubsection{HiddenKiller}
HiddenKiller is a stealthy backdoor attack that uses syntactic structures as triggers \cite{qi2021hidden}. The attack works by generating poisoned training samples through paraphrasing the clean dataset using a syntactically controlled model\textemdash SCPN \cite{iyyer2018adversarial}. The trigger pattern used is a low-frequency syntactic structure, \texttt{S(SBAR)(,)(NP)(VP)(.)}, which subtly alters sentence structure while preserving semantics \cite{qi2021hidden}. Each component corresponds to a syntactic unit: \texttt{S} is the full sentence, \texttt{SBAR} is a subordinate clause (e.g., "when..."), followed by a comma, a noun phrase (\texttt{NP}) as the subject, a verb phrase (\texttt{VP}) as the predicate, and a final period.

\subsubsection{StyleBkd}
StyleBkd is also a stealthy backdoor attack that uses text style transfer as triggers \cite{qi2021mind}. This attack modifies text using a pre-trained style transfer model, STRAP \cite{krishna2020reformulating}, which transforms the text to resemble the style of the Bible or poetry while preserving its semantic content. This attack method is highly invisible with a high attack success rate (ASR > 90\%) \cite{qi2021mind}, which shows strong resistance to defenses such as ONION\cite{qi2021onion}, PURE\cite{zhao2024defense}.

\subsection{Baseline Methods}
We evaluate the effectiveness of our approach against several established defense baselines \cite{zhao2024defense} designed to mitigate backdoor threats in transformer-based models. These include \textbf{Vanilla Fine-Tuning (FT)}, which applies standard fine-tuning without defenses~\cite{zhao2024defense}, and \textbf{Fine-Tuning with a Higher Learning Rate (FTH)}, which uses a rate of 5e-5 to potentially override poisoned weights~\cite{kurita2020weight}. \textbf{Maximum Entropy Fine-Tuning (MEFT)} introduces entropy regularization during early training to disrupt backdoor patterns~\cite{liu2023maximum}, followed by normal fine-tuning. We also compare against \textbf{PURE}, a variance-based method that prunes attention heads and applies attention normalization to suppress poisoned features~\cite{zhao2024defense}.

\subsection{Results and Analysis}
Table~\ref{tab:hiddenkiller} and Table~\ref{tab:stylebkd} present results on SST-2 under two types of backdoor attacks. For the syntactic trigger (Table~\ref{tab:hiddenkiller}), vanilla fine-tuning (FT) shows high clean accuracy (91.94\%) but a high label flip rate (LFR) of 41.73\%, indicating vulnerability to backdoor manipulation. Gradient-based pruning performs best, reducing the LFR to 31.71\% while preserving clean accuracy (91.61\%). When combined with structured L1/L2 sparsification, the method further boosts accuracy (92.69\%) and keeps LFR relatively low (33.62\%).  For the stylistic trigger (Table~\ref{tab:stylebkd}), increasing the learning rate (FTH) helps reduce LFR to 28.22\%, and PURE achieves similar results (LFR of 29.53\%). However, reinforcement learning-based pruning outperforms all others with the highest clean accuracy (92.83\%) and a low LFR (28.11\%). Bayesian pruning closely follows, achieving 92.59\% accuracy and 29.52\% LFR, showing a strong balance between robustness and performance.

\begin{table}[ht]
\centering
\small
\caption{Performance of defense methods against HiddenKiller backdoor attacks on SST-2.}
\begin{tabular}{lcc}
\toprule
\textbf{Method}                                & \textbf{ACC (\%)}        & \textbf{LFR (\%)}        \\
\midrule
FT (fine‑tune only)                            & $91.94 \pm 0.31$         & $41.73 \pm 3.97$         \\
FTH (higher LR)                                & $91.53 \pm 0.29$         & $33.35 \pm 3.86$         \\
MEFT (max‑entropy FT)                          & $91.42 \pm 0.43$         & $49.16 \pm 3.10$         \\
PURE                          & $91.55 \pm 0.33$           & $34.53 \pm 0.91$           \\

\midrule
Gradient‑based Pruning                         & $91.61 \pm 0.52$         & \textbf{31.71 $\pm$ 0.85}         \\
Layer‑Wise Pruning                             & $92.55 \pm 0.19$         & $37.35 \pm 0.78$         \\
Gradient‑based + Structured Sparsification      & $92.69 \pm 2.14$         & $33.62 \pm 1.90$         \\
Randomized Pruning + Ensemble                  & $92.42 \pm 0.43$         & $37.54 \pm 2.50$         \\
Reinforcement Learning‑Based Pruning           & $92.70 \pm 0.37$         & $35.54 \pm 1.99$         \\
Bayesian Pruning                               & $92.61 \pm 0.24$         & $37.37 \pm 1.37$         \\
\bottomrule
\end{tabular}
\label{tab:hiddenkiller}
\end{table}

\begin{table}[ht]
\centering
\small
\caption{Performance of defense methods against StyleBkd backdoor attacks on SST-2.}
\begin{tabular}{lcc}
\toprule
\textbf{Method}                                & \textbf{ACC (\%)}        & \textbf{LFR (\%)}        \\
\midrule
FT (fine‑tune only)                            & $92.26 \pm 0.37$         & $35.37 \pm 2.05$         \\
FTH (higher LR)                                & $91.29 \pm 0.12$         & $28.22 \pm 3.82$         \\
MEFT (max‑entropy FT)                          & $91.69 \pm 0.19$         & $29.77 \pm 5.59$         \\
PURE                           & $91.67 \pm 0.31$           & $29.53 \pm 2.16$           \\

\midrule
Gradient‑based Pruning                         & $91.32 \pm 0.53$         & $30.29 \pm 1.36$         \\
Layer‑Wise Pruning                             & $92.53 \pm 0.43$         & $33.01 \pm 2.39$         \\
Gradient‑based + Structured Sparsification      & $90.96 \pm 1.13$         & $31.56 \pm 2.85$         \\
Randomized Pruning + Ensemble                  & $92.60 \pm 0.49$         & $32.31 \pm 4.34$         \\
Reinforcement Learning‑Based Pruning           & $92.83 \pm 0.23$         & \textbf{28.11 $\pm$ 1.52}         \\
Bayesian Pruning                               & $92.59 \pm 0.41$         & $29.52 \pm 1.25$         \\
\bottomrule
\end{tabular}
\label{tab:stylebkd}
\end{table}

To understand the impact of gradient-based pruning, we use t-SNE to project [CLS] embeddings from clean test data into 2D space. In the HiddenKiller scenario (Figure 2a), the original model shows tight clusters influenced by the trigger, while the pruned model forms distinct, shifted clusters, indicating the successful removal of backdoor-related representations. Similarly, the choice of accuracy threshold ($\tau$) is crucial in pruning, as it balances ACC and LFR. Higher $\tau$ preserves accuracy but may miss triggers, while lower $\tau$ enables stronger pruning at the risk of reduced performance. Figure 2b shows the plot between LFR versus ACC for two attacks: HiddenKiller and StyleBkd, with different $\tau$ with gradient-based pruning. Reducing $\tau$ from 0.95 to 0.85 decreases LFR without a significant decrease in ACC; thus, $\tau = 0.85$ is optimal.

\begin{figure}[htbp]
  \centering
  \begin{subfigure}[b]{0.48\linewidth}
    \centering
    \includegraphics[width=\linewidth]{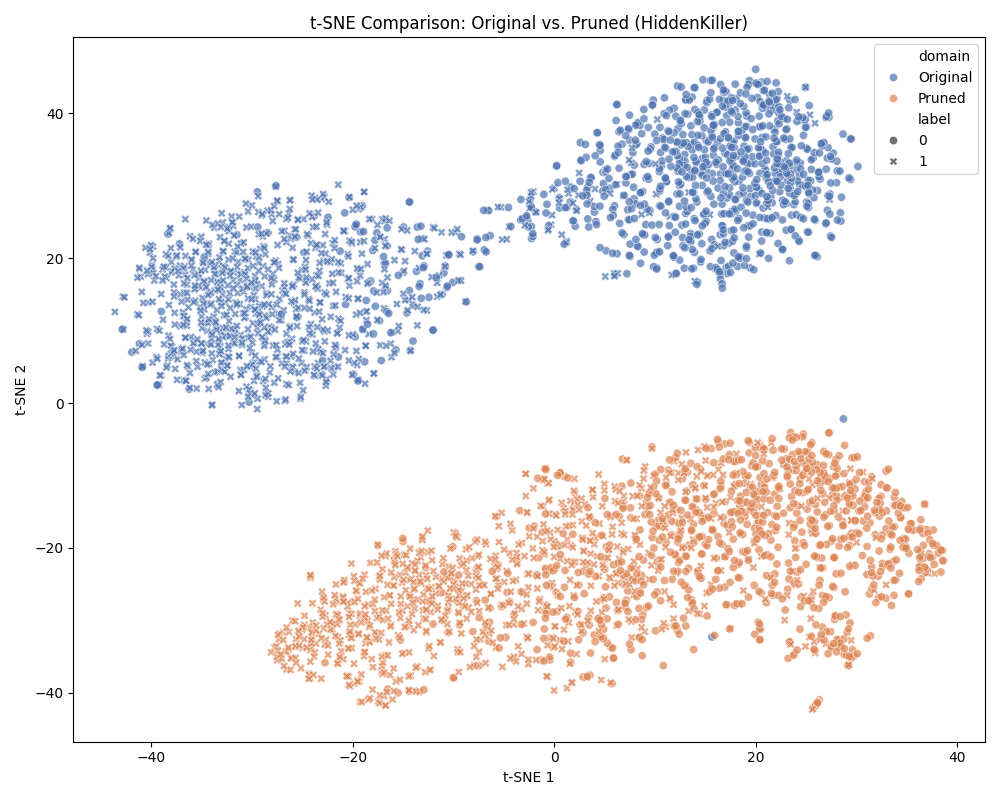}
    \caption{}
    \label{fig:tsnehk}
  \end{subfigure}
  \hfill
  \begin{subfigure}[b]{0.48\linewidth}
    \centering
    \includegraphics[width=\linewidth]{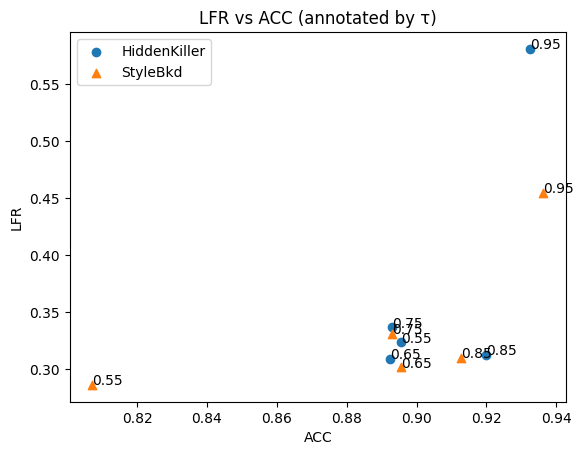}
    \caption{}
    \label{fig:tauplot}
  \end{subfigure}
  \caption{Visualization of embedding shift after gradient-based pruning and trade-off analysis for different $\tau$.}
  \label{fig:tsne_tau_combined}
\end{figure}

\section{Conclusion}
Our experiments show that pruning strategies are a possible defense method against backdoor attacks in transformer models, even when the end users lack the trigger knowledge or reference to an unpoisoned model. Among different evaluated models, gradient-based pruning achieved the best performance against syntactic backdoor attacks by reducing the LFR while maintaining clean accuracy. Future works could explore hybrid pruning. Another area could be developing an interactive visualization tool for monitoring the pruning process in real-time to better understand the model's vulnerabilities. At last, exploring such models in a multimodal transformer setting is another important step for better security across different NLP applications.


\begin{acks}
Shah Muhammad Hamdi is supported by the GEO directorate under NSF awards \#2301397 and \#2530946. Soukaina Filali Boubrahimi is supported by GEO Directorate under NSF awards \#2204363, \#2240022, and \#2530946.
\end{acks}

\section{GenAI Usage Disclosure}
Grammarly and ChatGPT-4, were used for grammatical refinement and language polishing.
\bibliographystyle{ACM-Reference-Format}
\bibliography{sample-base}

\appendix

\end{document}